\documentclass[11pt,a4paper]{article}
\usepackage[hyperref]{acl2019}
\usepackage{times}
\usepackage{latexsym}
\usepackage{graphicx}
\usepackage{graphics}
\usepackage{url}
\usepackage{amsmath}
\usepackage{url}
\usepackage{enumitem}
\usepackage{color}
\usepackage{booktabs}
\usepackage{multirow}
\aclfinalcopy 

\usepackage{todonotes}
\usepackage{ftnxtra}

\title{Emotional Neural Language Generation Grounded in Situational Contexts}

\author{Sashank Santhanam  and Samira Shaikh\\
  Department of Computer Science \\
  University of North Carolina at Charlotte \\
  Charlotte, NC, USA\\
  \texttt{\{ssantha1,samirashaikh\}@uncc.edu} \\}

\date{}

\begin{document}
\maketitle
\begin{abstract}
Emotional language generation is one of the keys to human-like artificial intelligence. Humans use different type of emotions depending on the situation of the conversation. Emotions also play an important role in mediating the engagement level with conversational partners. However, current conversational agents do not effectively account for emotional content in the language generation process. 
To address this problem, we develop a language modeling approach that generates affective content when the dialogue is situated in a given context. We use the recently released Empathetic-Dialogues corpus to build our models. Through detailed experiments, we find that our approach outperforms the state-of-the-art method on the perplexity metric by about 5 points and achieves a higher BLEU metric score. 
\end{abstract}

\section{Introduction}
Rapid advancement in the field of generative modeling through the use of neural networks has helped advance the creation of more intelligent conversational agents. Traditionally these conversational agents are built using \textit{seq2seq} framework that is widely used in the field of machine translation \cite{vinyals2015neural}. However, prior research has shown that engaging with these agents produces dull and generic responses whilst also being inconsistent with the emotional tone of conversation \cite{vinyals2015neural,li-etal-2016-persona}. These issues also affect engagement with the conversational agent, that leads to short conversations \cite{venkatesh2018evaluating}. Apart from producing engaging responses, understanding the situation and producing the right emotional response to a that situation is another desirable trait \cite{rashkin2019towards}. 

Emotions are intrinsic to humans and help in creation of a more engaging conversation \cite{poria2019emotion}. Recent work has focused on approaches towards incorporating emotion in conversational agents \cite{asghar2018affective,zhou2018emotional,huang2018automatic,ghosh-etal-2017-affect}, however these approaches are focused towards seq2seq task. We approach this problem of emotional generation as a form of transfer learning, using large pretrained language models. These language models, including BERT, GPT-2 and XL-Net, have helped achieve state of the art across several natural language understanding tasks \cite{devlin-etal-2019-bert,radford2019language,yang2019xlnet}. However, their success in language modeling tasks have been inconsistent \cite{ziegler2019encoder}. In our approach, we use these pretrained language models as the base model and perform transfer learning to fine-tune and condition these models on a given emotion. This helps towards producing more emotionally relevant responses for a given situation. In contrast, the work done by Rashkin \emph{et al.} \citeyearpar{rashkin2019towards} also uses large pretrained models but their approach is from the perspective of seq2seq task. 

Our work advances the field of conversational agents by applying the transfer learning approach towards generating emotionally relevant responses that is grounded on emotion and situational context. We find that our fine-tuning based approach outperforms the current state of the art approach on the automated metrics of the BLEU and perplexity. We also show that transfer learning approach helps produce well crafted responses on smaller dialogue corpus.

\section{Approach}
Consider the example show in Table \ref{problem_base} that shows a snippet of the conversation between a speaker and a listener that is grounded in a situation representing a type of emotion. Our goal is to produce responses to conversation that are emotionally appropriate to the situation and emotion portrayed.
\begin{table}[h]
\centering
\small
\begin{tabular}{@{}l@{}}
\toprule
\multicolumn{1}{l}{\textbf{Emotion: Confident}} \\ 
\multicolumn{1}{l}{\begin{tabular}[l]{@{}l@{}}\textbf{Situation:} I just knew I was going to do well at\\ work this morning.\end{tabular}} \\ \midrule
\begin{tabular}[c]{@{}l@{}}\textbf{Speaker:} I just knew I was going to do well at\\ work this morning. I was prepared\\ \textbf{Listener:} That is the way to go! Keep it up!\end{tabular} \\ \midrule
\end{tabular}
\caption{Example of conversations between a speaker and a listener}
\label{problem_base}
\end{table}
We approach this problem through a language modeling approach. We use large pre-trained language model 
as the base model for our response generation. This model is based on the transformer architecture and makes uses of the multi-headed self-attention mechanism to condition itself of the previously seen tokens to its left and produces a distribution over the target tokens. Our goal is to make the language model $p(y)=p(y_1,y_2,....,y_t;\theta)$ learn on new data and estimate the conditional probability $p(y|x)$. Radford \emph{et al.} 
\citeyearpar{radford2019language} demonstrated the effectiveness of language models to learn from a zero-shot approach in a multi-task setting. We take inspiration from this approach to condition our model on the task-specific variable $p(y_t|x,y_{< t})$, where $x$ is the task-specific variable, in this case the emotion label. We prepend the conditional variable (emotion, situational context) to the dialogue similar to the approach from Wolf \emph{et al} \citeyearpar{wolf2019transfertransfo}. We ensure that that the sequences are separated by special tokens.

\section{Experiments}
\subsection{Data}
In our experiments we use the Empathetic Dialogues dataset made available by Rashkin \emph{et al.} \citeyearpar{rashkin2019towards}. Empathetic dialogues is crowdsourced dataset that contains dialogue grounded in a emotional situation. The dataset comprises of 32  emotion labels including \textit{surprised, excited, angry, proud, grateful}. The speaker initiates the conversation using the grounded emotional situation and the listener responds in an appropriate manner\footnote{More information about the dataset made available on the \cite{rashkin2019towards}}.Table \ref{datasetstats} provides the basic statistics of the corpus. 
\begin{table}[htp]
\centering
\small
\begin{tabular}{cccc}
\toprule
\textbf{}           & \textbf{Train} & \textbf{Valid.} & \textbf{Test} \\
\toprule
\textbf{Num. Conversations}  & 19433           & 2770           & 2547          \\ \midrule
\textbf{Utterances} & 84324         & 12078           & 10973         \\ \midrule
\begin{tabular}[c]{@{}c@{}}\textbf{Avg Length}\\ \textbf{Conversations}\end{tabular}   & 4.31            & 4.36            & 4.31           \\ \midrule
\end{tabular}
\caption{Statistics of Empathetic Dialogue dataset used in our experiments}
\label{datasetstats}
\end{table}
\subsection{Implementation}
In all our experiments, we use the GPT-2 pretrained language model. We use the publicly available model containing 117M parameters with 12 layers; each layer has 12 heads. We implemented our models using PyTorch Transformers.\footnote{\url{https://github.com/huggingface/pytorch-transformers}} The input sentences are tokenized using byte-pair encoding(BPE) \cite{sennrich-etal-2016-neural} (vocabulary size of 50263). While decoding, we use the nucleus sampling ($p=0.9$) approach instead of beam-search to overcome the drawbacks of beam search \cite{holtzman2019curious,ippolito-etal-2019-comparison}. All our models are trained on a single TitanV GPU and takes around 2 hours to fine-tune the model. The fine-tuned models along with the configuration files and the code will be made available at: \url{https://github.com/sashank06/CCNLG-emotion}.

\subsection{Metrics}
Evaluating the quality of responses in open domain situations where the goal is not defined is an important area of research. Researchers have used methods such as BLEU , METEOR \cite{banerjee2005meteor}, ROUGE \cite{lin2004rouge} from machine translation and text summarization \cite{dialogue-eval} tasks. BLEU and METEOR are based on word overlap between the proposed and ground truth responses; they do not adequately account for the diversity of responses that are possible for a given input utterance and show little to no correlation with human judgments \cite{dialogue-eval}. We report on the BLEU \cite{papineni2002bleu} and Perplexity (PPL) metric to provide a comparison with the current state-of-the-art methods. We also report our performance using other metrics such as length of responses produced by the model. Following, Mei \emph{et al} \citeyearpar{mei17}, we also report the diversity metric that helps us measure the ability of the model to promote diversity in responses \cite{diversity2016}. Diversity is calculated as the as the number of distinct unigrams in the generation scaled by the total number of generated tokens \cite{mei17,li-etal-2016-persona}. We report on two additional automated metrics of readability and coherence. Readability quantifies the linguistic quality of text and the difficulty of the reader in understanding the text \cite{novikova-etal-2017-need}. We measure readability through the Flesch Reading Ease (FRE) \cite{kincaid1975derivation} which computes the number of words, syllables and sentences in the text. Higher readability scores indicate that utterance is easier to read and comprehend. Similarly, coherence measures the ability of the dialogue system to produce responses consistent with the topic of conversation. To calculate coherence, we use the method proposed by Dziri \emph{et al.} \citeyearpar{dziri2018augmenting}.

\section{Results}
\subsection{Automated Metrics}
\begin{table*}[h]
\centering
\small
\begin{tabular}{ccccccc}
\toprule
\textbf{Experiment}                                              & \textbf{\begin{tabular}[c]{@{}c@{}}Valid\\ PPL\end{tabular}} & \textbf{BLEU} & \textbf{Readability} & \textbf{Coherence} & \textbf{Length} & \textbf{Diversity} \\ \toprule
\begin{tabular}[c]{@{}c@{}}Baseline\\ Fine-Tuned \\ \cite{rashkin2019towards} \end{tabular}    & 21.24                                                        & 6.27          & x                    & x                  & x               & x                  \\ \midrule
\begin{tabular}[c]{@{}c@{}}Baseline\\ Emo-prepend \\ \cite{rashkin2019towards}\end{tabular}  & 24.30                                                        & 4.36          & x                    & x                  & x               & x                  \\ \midrule
\begin{tabular}[c]{@{}c@{}}Our Model\\ Fine-Tuned\end{tabular}   & \textbf{18.32}                                                        & 7.71          & 0.78                 & 0.93               & 9.77            & 0.0031             \\ \midrule
\begin{tabular}[c]{@{}c@{}}Our Model\\ Emo-prepend\end{tabular} & 19.49                                                        & \textbf{7.78}          & \textbf{0.79}                 & 0.93               & 9.71            & \textbf{0.0033}      \\ \bottomrule      
\end{tabular}
\caption{Comparison of the performance of our model to the baseline model proposed by Rashkin \emph{et al} \citeyearpar{rashkin2019towards} across a variety of automated metrics to provide a thorough comparison. \textbf{x} indicates that these metrics were not provided in the Rashkin \emph{et al} \citeyearpar{rashkin2019towards} work. }
\label{result_automated}
\end{table*}
We first compare the performance of our approach with the baseline results obtained from Rashkin \emph{et al.} \citeyearpar{rashkin2019towards} that uses a full transformer architecture \cite{vaswani2017attention}, consisting of an encoder and decoder. Table \ref{result_automated} provides a comparison of our approach with to the baseline approach. In Table \ref{result_automated}, we refer our ``\textit{Our Model Fine-Tuned}'' as the baseline fine-tuned GPT-2 model trained on the dialogue and ``\textit{Our-model Emo-prepend}'' as the GPT-2 model that is fine-tuned on the dialogues but also conditioned on the emotion displayed in the conversation. We find that fine-tuning the GPT-2 language model using a transfer learning approach helps us achieve a lower perplexity and a higher BLEU scores. The results from our approach are consistent with the empirical study conducted by Edunov \emph{et al} \citeyearpar{edunov-etal-2019-pre} that demonstrate the effectiveness of the using pre-trained model diminishes when added to the decoder network in an \textit{seq2seq} approach. We also perform a comparison between our two models on the metrics of length, diversity, readability and coherence. We find that our baseline model produces less diverse responses compared to when the model is conditioned on emotion. We find that the our \textit{emo-prepend} model also higher a slightly higher readability score that our baseline model.

\subsection{Qualitative Evaluation}
To assess the quality of generations, we conducted a MTurk human evaluation. We recruited a total of 15 participants and each participant was asked to evaluate 25 randomly sampled outputs from the test set on three metrics:
\begin{enumerate}[noitemsep,nolistsep]
    \item Readability - Is the response easy to understand, fluent and grammatical and does not have any consecutive repeating words.
    \item Coherence - Is the response relevant to the context of the conversation.
    \item Emotional Appropriateness- Does the response convey emotion suitable to the context of the conversation?
\end{enumerate}
\begin{table*}[h]
\centering
\small
\begin{tabular}{lll}
\toprule
\multicolumn{1}{c}{\textbf{\begin{tabular}[c]{@{}c@{}}Input \\ Conversation\end{tabular}}}                                                           & \multicolumn{1}{c}{\textbf{Model}}                     & \multicolumn{1}{c}{\textbf{Outputs}}                                                                     \\ \toprule
\multirow{3}{*}{\begin{tabular}[c]{@{}l@{}}Person A: I think I aced my  supervisor \\ exam!\end{tabular}}                                             & \begin{tabular}[c]{@{}l@{}}Ground\\ Truth\end{tabular} & \begin{tabular}[c]{@{}l@{}}Woo hoo! That is awesome! Congrats! When will you find out\end{tabular} \\ \cmidrule(l){2-3}
                                                                                                                                                     & \begin{tabular}[c]{@{}l@{}}Fine\\ Tuned\end{tabular}   & What was it?                                                                                             \\\cmidrule(l){2-3}
                                                                                                                                                     & \begin{tabular}[c]{@{}l@{}}Emo\\ Prepend\end{tabular}  & \begin{tabular}[c]{@{}l@{}}Wow, that's  impressive!\end{tabular}                               \\ \midrule
\multirow{3}{*}{\begin{tabular}[c]{@{}l@{}}Person A: I had pizza at a restaurant a\\ couple days ago and found  a hair on my \\pizza!\end{tabular}} & \begin{tabular}[c]{@{}l@{}}Ground\\ Truth\end{tabular} & \begin{tabular}[c]{@{}l@{}}It is disgusting. What did  you do?\end{tabular}                            \\ \cmidrule(l){2-3}
                                                                                                                                                     & \begin{tabular}[c]{@{}l@{}}Fine\\ Tuned\end{tabular}   & I was so angry                                                                                           \\ \cmidrule(l){2-3}
                                                                                                                                                     & \begin{tabular}[c]{@{}l@{}}Emo\\ Prepend\end{tabular}  & \begin{tabular}[c]{@{}l@{}}That's terrible,  did you clean it up?\end{tabular}        \\ \bottomrule         
\end{tabular}
\label{samples}
\caption{Example generations from our two model along with the ground truth responses. }
\end{table*}
\begin{table}[h]
\centering
\small
\begin{tabular}{p{1.6cm}p{1.4cm}p{1.4cm}p{1.5cm}}
\toprule
\textbf{}           & \textbf{Readability} & \textbf{Coherence} & \begin{tabular}[c]{@{}c@{}}\textbf{Emotional} \\ \textbf{Appropriate}-\\ \textbf{ness} \end{tabular} \\
\toprule
\begin{tabular}[c]{@{}c@{}}Our Model\\ Fine-Tuned\end{tabular} & 4.14           & 3.50           & 3.70          \\ \midrule
\begin{tabular}[c]{@{}c@{}}Our Model\\ Emo-prepend\end{tabular} & 3.54         & 3.4           & 3.19         \\ \midrule
\begin{tabular}[c]{@{}c@{}}Ground 
\\Truth \end{tabular}   & 3.92           & 3.86            & 4           \\ \midrule
\end{tabular}
\caption{Human ratings demonstrating a comparison between our models to the ground truth responses on the metrics of readability, coherence and emotional appropriateness}
\label{humaneval}
\end{table}
Table \ref{humaneval} shows the results obtained from the human evaluation comparing the performance of our fine-tuned, emotion pre-pend model to the ground-truth response. We find that our fine-tuned model outperforms the emo-prepend on all three metrics from the ratings provided by the human ratings.


\section{Related Work}
The area of dialogue systems has been studied extensively in both open-domain \cite{niu2018polite} and goal-oriented \cite{lipton2018bbq} situations. Extant approaches towards building dialogue systems has been done predominantly through the \textit{seq2seq} framework \cite{vinyals2015neural}. However, prior research has shown that these systems are prone to producing dull and generic responses that causes engagement with the human to be affected \cite{vinyals2015neural,venkatesh2018evaluating}. Researchers have tackled this problem of dull and generic responses through different optimization function such as MMI \cite{li-etal-2016-diversity} and through reinforcement learning approaches\cite{Li-reinforce}. Alternative approaches towards generating more engaging responses is by grounding them in personality of the speakers that enables in creating more personalized and consistent responses \cite{li-etal-2016-persona,zhang-etal-2018-personalizing,wolf2019transfertransfo}.

Several other works have focused on creating more engaging responses by producing affective responses. One of the earlier works to incorporate affect through language modeling is the work done by Ghosh 
\emph{et al.} \cite{ghosh-etal-2017-affect}. This work leverages the LIWC \cite{pennebaker2001linguistic} text analysis platform for affective features. Alternative approaches of inducing emotion in generated responses from a \textit{seq2seq} framework include the work done by Zhou \emph{et al}\citeyearpar{zhou2018emotional} that uses internal and external memory, Asghar \emph{et al.} \citeyearpar{asghar2018affective} that models emotion through affective embeddings and Huang 
\emph{et al} \citeyearpar{huang2018automatic} that induce emotion through concatenation with input sequence. More recently, introduction of transformer based approaches have helped advance the state of art across several natural language understanding tasks \cite{vaswani2017attention}. These transformers models have also helped created large pre-trained language models such as BERT \cite{devlin-etal-2019-bert}, XL-NET \cite{yang2019xlnet}, GPT-2 \cite{radford2019language}. However, these pre-trained models show inconsistent behavior towards language generation \cite{ziegler2019encoder}. 
\section{Conclusion and Discussion}
In this work, we study how pre-trained language models can be adopted for conditional language generation on smaller datasets. Specifically, we look at conditioning the pre-trained model on the emotion of the situation produce more affective responses that are appropriate for a particular situation. We notice that our fine-tuned and emo-prepend models outperform the current state of the art approach relative to the automated metrics such as BLEU and perplexity on the validation set. We also notice that the emo-prepend approach does not out perform a simple fine tuning approach on the dataset. We plan to investigate the cause of this in future work from the perspective of better experiment design for evaluation \cite{santhanam2019towards} and analyzing the models focus when emotion is prepended to the sequence \cite{clark2019does}. Along with this, we also notice other drawbacks in our work such as not having an emotional classifier to predict the outcome of the generated sentence, which we plan to address in future work. 
\label{conclusion}

 \section*{Acknowledgments}
This work was supported by the Defense Advanced Research Projects Agency (DARPA) under Contract No FA8650-18-C-7881. All statements of fact, opinion or conclusions contained herein are those of the authors and should not be construed as representing the official views or policies of AFRL, DARPA, or the U.S. Government. We thank the anonymous reviewers for the helpful feedback. 

\bibliography{acl2019}
\bibliographystyle{acl_natbib}

\end{document}